\documentclass{article}
\usepackage{ijcai17}
\usepackage{times}
\usepackage[utf8]{inputenc}
\usepackage{mathtools}
\usepackage{graphicx}
\usepackage{caption}
\usepackage{url}
\usepackage{subcaption}
\usepackage[linesnumbered]{algorithm2e}
\usepackage{amsfonts}
\usepackage{amssymb}
\usepackage{dsfont}
\renewcommand{\Re}{\mathds{R}}

\usepackage{xcolor}

\usepackage[normalem]{ulem}

\title{A method for the online construction of the set of states of a Markov Decision Process using Answer Set Programming}

 \author{Leonardo A. Ferreira\(^1\),  Reinaldo A. C. Bianchi\(^2\),  Paulo E. Santos\(^2\), Ramon Lopez de Mantaras\(^3\)\\
 \(^1\)Universidade Metodista de São Paulo,  S\~ao Bernardo do Campo, Brazil.  \\
 \(^2\)Centro Universitário FEI, S\~ao Bernardo do Campo, Brazil.  \\
 \(^3\)IIIA-CSIC, Bellaterra, Espa\~na.\\
 leonardo.ferreira@metodista.br, \{rbianchi,psantos\}@fei.edu.br, mantaras@iiia.csic.es}

\begin{document}
\maketitle
\begin{abstract}
    Non-stationary domains, that change in unpredicted ways, are a challenge for agents searching for optimal policies in sequential decision-making problems. This paper presents a combination of Markov Decision Processes (MDP) with Answer Set Programming (ASP), named {\em Online ASP for MDP} (oASP(MDP)), which is a method capable of constructing the set of domain states while the agent interacts with a changing environment. oASP(MDP) updates previously obtained policies, learnt by means of Reinforcement Learning (RL), using rules that represent the domain changes observed by the agent. These rules represent a set of domain constraints that are processed as ASP programs reducing the search space. Results show that oASP(MDP) is capable of finding solutions for problems in non-stationary domains without interfering with the action-value function approximation process.
\end{abstract}
\section{Introduction}\label{sec:intro}
A key issue in Artificial Intelligence (AI) is to equip autonomous agents with the ability to operate in changing domains by adapting the agents' processes at a cost that is equivalent to the complexity of the domain changes. This ability is called {\em elaboration tolerance}~\cite{McCarthy1987,McCarthy98}. Consider, for instance, an autonomous robot learning to navigate in an unknown environment. Unforeseen events may happen that could block  passages (or open previously unavailable ones). The autonomous agent should be able to find new solutions in this changed domain using the knowledge previously acquired plus the knowledge acquired from the observed changes in the environment, without having to operate a complete code-rewriting, or start a new cycle of domain-exploration from scratch. 
 
Reinforcement Learning (RL) is an AI framework in which an agent interacts with its environment in order to find a sequence of actions (a policy) to perform a given task \cite{Sutton-Barto2ed}. RL is capable of finding optimal solutions to Markov Decision Processes (MDP) without assuming total information about the problem's domain. However, in spite of having the optimal solution to a particular task, a RL agent may still perform poorly on a new task, even if the latter is similar to the former \cite{Shanahan2016}. Therefore, Reinforcement Learning alone does not provide elaboration-tolerant solutions. Non-monotonic reasoning can be used as a tool to increase the generality of domain representations \cite{McCarthy1987} and may provide the appropriate element to build agents more adaptable to changing situations. In this work we consider Answer Set Programming (ASP) \cite{Gelfond88,Lifschitz2002}, which is a declarative non-monotonic logic programming language, to bridge the gap between RL and elaboration tolerant solutions. The present paper tackles this problem by introducing a novel algorithm: {\em Online ASP for MDP} (oASP(MDP)), that updates previously obtained policies, learned by means of Reinforcement Learning (RL), using rules that represent the domain changes as observed by the agent. These rules are constructed by the agent in an {\em online} fashion (i.e., as the agent perceives the changes) and they impose constraints on the domain states that are further processed by an ASP engine, reducing the search space. Tests performed in non-stationary non-deterministic grid worlds show that, not only oASP(MDP) is capable of finding the action-value function for an RL agent and, consequently, the optimal solution, but also that using ASP does not hinder the performance of a learning agent and can improve the overall agent's performance.

To model an oASP(MDP) learning agent (Section~\ref{sec:prop}), we propose the combination of Markov Decision Processes and Reinforcement Learning (Section~\ref{sec:mdp}) with ASP (Section~\ref{sec:asp}). Tests were performed in two different non-stationary non-deterministic grid worlds (Section~\ref{sec:exp}), whose results show a considerable increase in the agent's performances when compared with a RL base algorithm, as presented in Sections~\ref{sec:exp1} 
and~\ref{sec:exp3}.
\section{Background}\label{sec:bg}
This section introduces Markov Decision Processes (MDP), Reinforcement Learning (RL) and Answer Set Programming (ASP) that are the foundations of the work reported in this paper.
\subsection{MDP and Reinforcement Learning}\label{sec:mdp}
In a sequential decision making problem, an agent is required to execute a series of actions in an environment in order to find the solution of a given problem. Such sequence of actions, that forms a feasible solution, is known as a policy (\(\pi\)) which leads the agent from an initial state to a goal state \cite{Bellman1957,Bellman-Dreyfus1962}. Given a set of feasible solutions, an optimal policy \(\pi^\ast\) can be found by using Bellman's Principle of Optimality~\cite{Bellman-Dreyfus1962}, which states that ``an optimal policy has the property that whatever the initial state and initial decision are, the remaining decisions must constitute an optimal policy with regard to the state resulting from the first decision''; \(\pi^\ast\) can be defined as the policy that maximises/minimises a desired reward/cost function.

A formalisation that can be used to describe sequential decision making problems is a Markov Decision Process (MDP) that is defined as a tuple \(\langle \mathcal{S}, \mathcal{A}, \mathcal{T}, \mathcal{R} \rangle\), where:
\begin{itemize}
    \item \(\mathcal{S}\) is the set of states that can be observed in the domain;
    \item \(\mathcal{A}\) is the set of actions that the agent can execute;
    \item \(\mathcal{T} : \mathcal{S} \times \mathcal{A} \times \mathcal{S} \mapsto [0,1]\) is the transition function that provides the probability of, being in \(s \in \mathcal{S}\) and executing \(a \in \mathcal{A}\), reaching the future state \(s' \in \mathcal{S}\);
    \item \(\mathcal{R} : \mathcal{S} \times \mathcal{A} \times \mathcal{S} \mapsto \Re\) is the reward function that provides a real number when executing \(a \in \mathcal{A}\) in the state \(s \in \mathcal{S}\) and observing \(s' \in \mathcal{S}\) as the future state.
\end{itemize}

One method that can be used to find an optimal policy for MDPs, which does not need \emph{a priori} knowledge of the transition and reward functions, is the reinforcement learning model-free off-policy method known as Q-Learning~\cite{Watkins,Sutton-Barto2ed}. 

Given an MDP \(\mathcal{M}\), Q-Learning learns while an agent interacts with its environment by executing an action \(a_t\) in the current state \(s_t\) and observing both the future state \(s_{t+1}\) and the reward \(r_{t+1}\). With these observations, Q-Learning updates an action-value function \(Q(s,a)\) using
{\small{\[Q(s_t,a_t) \gets Q(s_t,a_t) + \alpha . (r_{t+1} + \gamma . \max_a Q(s_{t+1},a) - Q(s_t,a_t))\mathrm{,}\]}}
where \(\alpha\) is the learning rate and \(\gamma\) is a discount factor. By using these reward values to approximate a \(Q(s,a)\) function that maps a real value to pairs of states and actions, Q-Learning is capable of finding \(\pi^\ast\) which maximises the reward function. Since Q-Learning is a well-known and largely used RL method, we omit its detailed description here, which can be found in \cite{Watkins,Sutton-Barto2ed}.

Although Q-Learning does not need information about \(\mathcal{T}\) and \(\mathcal{R}\), it still needs to know the set \(\mathcal{S}\) of states before starting the interaction with the environment. For finding this set, this work uses Answer Set Programming.
\subsection{Answer Set Programming}\label{sec:asp}
Answer Set Programming (ASP) is a declarative non-monotonic logic programming language that has been successfully used for NP-complete problems such as planning~\cite{Lifschitz2002,Mohan2015IEEE,yang2014}.

An ASP rule is represented as 
\begin{equation}
A \gets L_1, L_2, \ldots, L_n \label{eq:rule}
\end{equation}
where \(A\) is an atom (the head of the rule) and the conjunction of  literals \(L_1, L_2, \ldots, L_n\) is the rule's body.

An ASP program \(\Pi\) is a set of rules in the form of Formula~\ref{eq:rule}. ASP is based on the stable model semantics of logic programs \cite{Gelfond08}. A stable model of \(\Pi\) is an interpretation that makes every rule in \(\Pi\) true, and is a minimal model of \(\Pi\). ASP programs are executed by computing stable models, which is usually accomplished by inference engines called answer set solvers \cite{Gelfond08}.



Two important aspects of ASP are its third truth value for \emph{unkown}, along with \emph{true} and \emph{false}, and its two types of negation:  strong (or classical) negation and weak negation, representing \emph{negation as failure}. As it is defined over stable models semantics, ASP respects the rationality that \emph{one shall not believe anything one is not forced to believe} \cite{Gelfond88}.


Although ASP does not allow explicit reasoning with or about probabilities, ASP's choice rules are capable of generating distinct outcomes for the same input. I.e., given a current state \(s\) and an action \(a\), it is possible to describe in an ASP logic program states \(s1\), \(s2\) and \(s3\) as possible outcomes of executing \(a\) in \(s\) as ``\texttt{1\{ s1, s2, s3 \}1 :- a, s.}''. Such choice rules can be read as ``given that \(s\) and \(a\) are true, choose at least one and at maximum of one state from \(s1\), \(s2\) and \(s3\)''. Thus,  the answer sets \texttt{[s, a, s1]}, \texttt{[s, a, s2]} and \texttt{[s, a, s3]} represent the possible transitions that are the effects of executing action \(a\) on state \(s\).

This work assumes that for each state \(s \in \mathcal{S}\) there is an ASP logic program with choice rules describing the consequences of each action \(a \in \mathcal{A}_s\)  (where \(\mathcal{A}_s\subseteq \mathcal{A}\) is the set of actions for the state $s$). ASP programs can also be used to represent domain constraints: the allowed or forbidden states or actions. In this context, to find a set \(\mathcal{S}\) of an MDP and its \(Q(s,a)\) function is to find every answer set for every state that the agent is allowed to visit, i.e. every allowed transition for each state-action pair. In this paper ASP is used to find the set of states \(\mathcal{S}\) of an MDP and Q-Learning is used to approximate \(Q(s,a)\) without assuming prior knowledge of \(\mathcal{T}\) and \(\mathcal{R}\). The next section describes this idea in more details.

\section{Online ASP for MDP: oASP(MDP)}\label{sec:prop}
Given sets \(\mathcal{S}\) and \(\mathcal{A}\) of an MDP, a RL method \(M\) can approximate an action-value function \(Q(s,a)\). If \(\mathcal{S}\) is constructed state by state while the agent is interacting with the world, \(M\) is still able to approximate \(Q(s,a)\), as it only uses the current and past states for that. By using choice rules in ASP, it is possible to describe a transition \(t(s,a,s')\) in the form \texttt{1\{s'\}1 :- a} for each action \(a \in \mathcal{A}_s\) and each state \(s \in \mathcal{S}\). By describing possible transitions for each action in each state as a logic program, an ASP engine can be used to provide a set of observed states \(\mathcal{S}_o\), a set of actions \(\mathcal{A}_s\) for each state and, finally, an action-value function defined from the interaction with the environment, that can be used to further operate in this environment. This is the essence of the oASP(MDP) method, represented in Algorithm \ref{alg:oaspmdp}.

\begin{algorithm}[ht!]
    \TitleOfAlgo{oASP(MDP)}
    \BlankLine
    \KwIn{The set of actions \(\mathcal{A}\), an action-value function approximation method \(M\) and a number of episodes \(n\).}
    \KwOut{The approximated \(Q(s,a)\) function.}
\DontPrintSemicolon
    \BlankLine
    Initialize the set of observed states \(\mathcal{S}_o = \varnothing\)\;
    \While{number of episodes performed is less than \(n\)}{
        \Repeat{the end of the episode}{
            Observe the current state \(s\)\;
                \eIf{\(s \not \in \mathcal{S}_o\)}{
                    Add \(s\) to the set of states \(\mathcal{S}_o\).\;\label{alg:ifline}
                    Choose and execute a random action \(a \in \mathcal{A}\).\;
                    Observe the future state \(s'\).\;
                    Update state \(s\) logic program with observed transition adding a choice rule.\;
                    Update \(Q(s,a)\)'s description by finding every answer set for each state \(s\) added to \(\mathcal{S}_o\) in this episode.\;
                }{\label{alg:elseline}
                    Choose an action \(a \in \mathcal{A}\) as defined by \(M\).\;
                    Execute the chosen action \(a\).\;
                    Observe the future state \(s'\).\;
                }
            Update \(Q(s,a)\)'s value as defined by \(M\).\;\label{alg:update}
            Update the current state \(s \gets s'\).\;
                
    }
}
    \caption{The oASP(MDP) Algorithm.}
\label{alg:oaspmdp}
\end{algorithm}

\begin{figure}[t!]
    \centering
    \includegraphics[width=0.75\columnwidth]{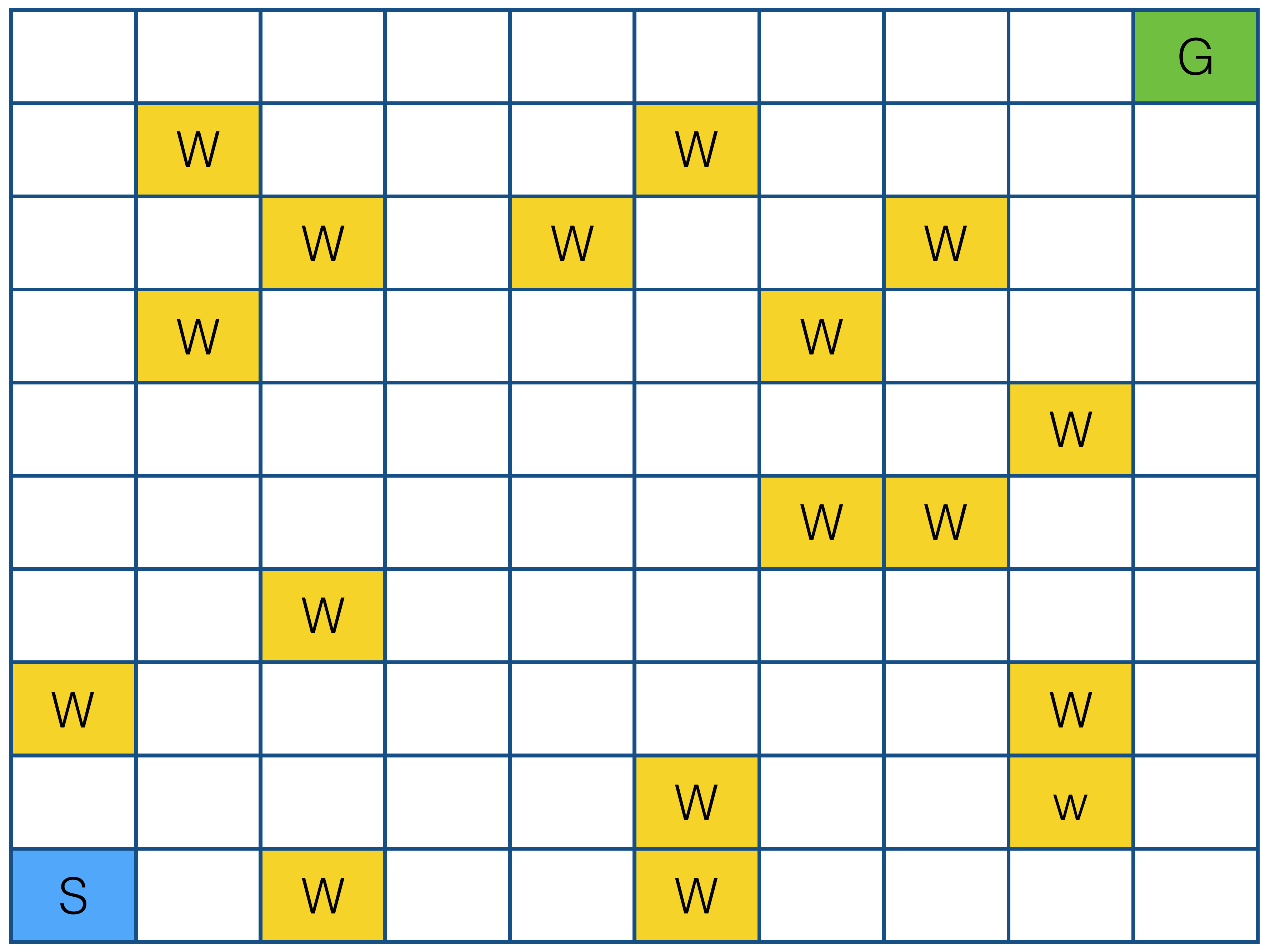}
    \caption{Example of a randomly generated grid world.}
    \label{fig:gridex}
\end{figure}

In order to illustrate oASP(MDP) (Algorithm~\ref{alg:oaspmdp}), let's consider the grid world in Figure~\ref{fig:gridex}, and an oASP(MDP) agent, initially located at the state ``S" (blue cell in the grid), that is capable of executing any action in the following set: \(\mathcal{A} = \)\{ \textit{go up}, \textit{go down}, \textit{go left}, \textit{go right}\}. This grid world has walls (represented by the letter ``W"), that are cells where the agent cannot occupy and through which it is unable to pass. If an agent moves toward a wall (or toward an external border of the grid) it stays at its original location. When the interaction with the environment starts, the agent has only information about the set of actions \(\mathcal{A}\). The set of observed states \(\mathcal{S}_o\) is initially empty. 

At the beginning of the agent's interactions with the environment, the agent observes the initial state \(s0\) and verifies if it is in \(\mathcal{S}_o\). Since \(s0 \not \in \mathcal{S}_o\), the agent adds \(s0\) to \(\mathcal{S}_o\) (line~\ref{alg:ifline} of Algorithm~\ref{alg:oaspmdp}) and executes a random action, let this action be \textit{go up}. As a consequence of this choice, the agent moves to a new state \(s1\) (the cell above S) and receives a reward \(r0\). At this moment, the agent has information about the previous state, allowing it to write the choice rule ``\(1\{s1\}1 :- \, s0, go\; up\)'' as an ASP logic program. In this first interaction, the only answer set that can be found for this choice rule is ``\([s0, go\; up, s1]\)''. With this information the agent can initialize a \(Q(s0,go \;up)\) and update this value using the reward \(r0\) (line~\ref{alg:update}).

After this first interaction, the agent is in the state \(s1\) (the cell above S). Again, this is an unknown state (\(s1 \not \in \mathcal{S}_o\)), thus, as with the previous state, the agent adds \(s1\) to \(\mathcal{S}_o\), chooses a random action, let it be \textit{go up} again, and executes this action in the environment. By performing \textit{go up} in this state, the agent hits a wall and stays in the same state. With this observation, the agent writes the choice rule ``\(1\{s1\}1 :-\, s1, go\, up\)'' and updates the value of \(Q(s1, go\;up)\) using the received reward \(r1\). 

Since the agent is in the same state as in the previous interaction, it knows the consequence of the action \textit{go up} in this state, but has no information about any other actions for this state. At this moment, the agent selects an action using the action-selection function defined by the learning method \(M\) and executes it in the environment. For example, let it choose \textit{go down}, returning to the blue cell (S). The state \(s1\) has now two choice rules: ``\(1\{s1\}1 :- \,s1, go\; up\)'' and ``\(1\{s0\}1 :- \,s1, go\; down\)'' which lead to the answer sets ``\([s1, go\; up]\)'' and ``\([s1, go\; down, s0]\)'' respectively. Once again, the agent updates the \(Q(s1,go\; up)\) function using the method described in \(M\) with the reward \(r2\) received. After this transition, the agent finds itself once again in the initial state and continues the domain exploration just described. If, for example, the agent chooses to execute the action \textit{go up} again, but due to the non-deterministic nature of the environment, the agent goes to the state on the right of the blue square, then a new state \(s2\) is observed and the choice rule for the previous state is updated to ``\(1\{s1, s2\}1 :- \,s0, go\; up\)''. The answer sets that can be found considering this choice rule are ``\([s0, go\; up, s1]\)'' and ``\([s0, go\; up, s2]\)''. With the reward \texttt{r3} received, the agent updates the value of \(Q(s0, go \;up)\).

The learning process of oASP(MDP) continues according to the chosen action-value function approximation method (from line~\ref{alg:elseline} onwards). After a number of interactions with the environment, the oASP(MDP) agent has executed every possible action in every state that \textit{is possible to be visited} and has the complete environment description. Note that this method excludes states of the MDP that are unreachable by the agent, which improves the efficiency of a RL agent in cases that the environment imposes state constrains (as we shall see in the next section).

The next section presents the tests applied to evaluate oASP(MDP) implemented with Q-Learning as the action-value function approximation method \(M\).

\section{Tests and Results}\label{sec:exp}

The oASP(MDP) algorithm was evaluated with tests performed in non-deterministic, non-stationary, grid-world domains. Two test sets were considered where, in each set, one of the following domain variables was randomly changed: {\em the number and location of walls in the grid} (first test, Section \ref{sec:exp1}), and {\em the transition probabilities} (second test, Section \ref{sec:exp3}). 

 Four actions were allowed in the test domains considered in this work: \textit{go up}, \textit{go down}, \textit{go left} and \textit{go right}. Each action has a predefined probability of conducting the agent in the desired direction and also for moving the agent to an orthogonal (undesired) location. The transition probability for each action depends on the grid world and will be defined for each test, as described below. In all tests, the initial state was fixed at the lower-leftmost square (e.g., cell `S' in Fig. \ref{fig:gridex}) and the goal state fixed in the upper-rightmost square (e.g., cell `G' in Fig. \ref{fig:gridex}). 
 
In the test domains, walls were distributed randomly in the grid as obstacles. For each grid, the ratio of walls per grid size is defined. The initial and goal states are the only cells that do not accept obstacles. Wall's placement in the grid changed at the 1000$^{th}$ and 2000$^{th}$ episodes during each test trial. An example of a grid used in this work is shown in Figure~\ref{fig:gridex}.

Results show the data obtained from executing Q-Learning and oASP(MDP) (with Q-Learning as the action-value function approximation method) in the same environment configuration. 
The values used for the learning variables were: learning rate \(\alpha = 0.2\), discount factor \(\gamma = 0.9\), exploration/exploitation rate for the \(\epsilon\)-greedy action selection method: \(\epsilon = 0.1\) and the maximum number of steps before an episode is finished was 1000.

In each test, three variables were used to compare Q-Learning and oASP(MDP). First, the root-mean-square deviation (RMSD), that provides information related to the convergence of the methods by comparing values of the \(Q(s,a)\) function in the current episode with respect to that obtained in the previous episode. Second, we considered the return (sum of the rewards) received in an episode. Third, the number of steps needed to go from the initial state to the goal state was evaluated. The results obtained were also compared with that of an agent using the optimal policy in a deterministic grid world (the best performance possible, shown as a red-dashed line in the results below).

For oASP(MDP), the number of state-action pairs known by the agent was also measured and compared with the size of Q-Learing's fixed \(Q(s,a)\) tabular implementation. This variable provides information of how far an oASP(MDP) agent is from knowing the complete environment along with how much the \(Q(s,a)\) function could be reduced.

The test domains and related results are described in details in the next sections.

\subsection{First test: changes in the wall--free-space ratio}\label{sec:exp1}

In the first test, the size of the grid was fixed to 10$\times$10 and the transition probabilities were assigned at 90\% for moving on the desired direction and 5\% for moving in each of the two directions that are orthogonal to the desired. In this test, changes in the environment occurred in the number and location of walls in the grid. Initially the domain starts with no walls (0\%), then it changes to a world where 10\% of the grid is occupied by walls placed at random locations and, finally, the grid world changes to a situation where 25\% of the grid is occupied by walls. Each change occurs after 1000 episodes.

\begin{figure}[ht!]
\centering
\begin{subfigure}[t]{0.85\columnwidth}
	\centering
	\includegraphics[width=\columnwidth]{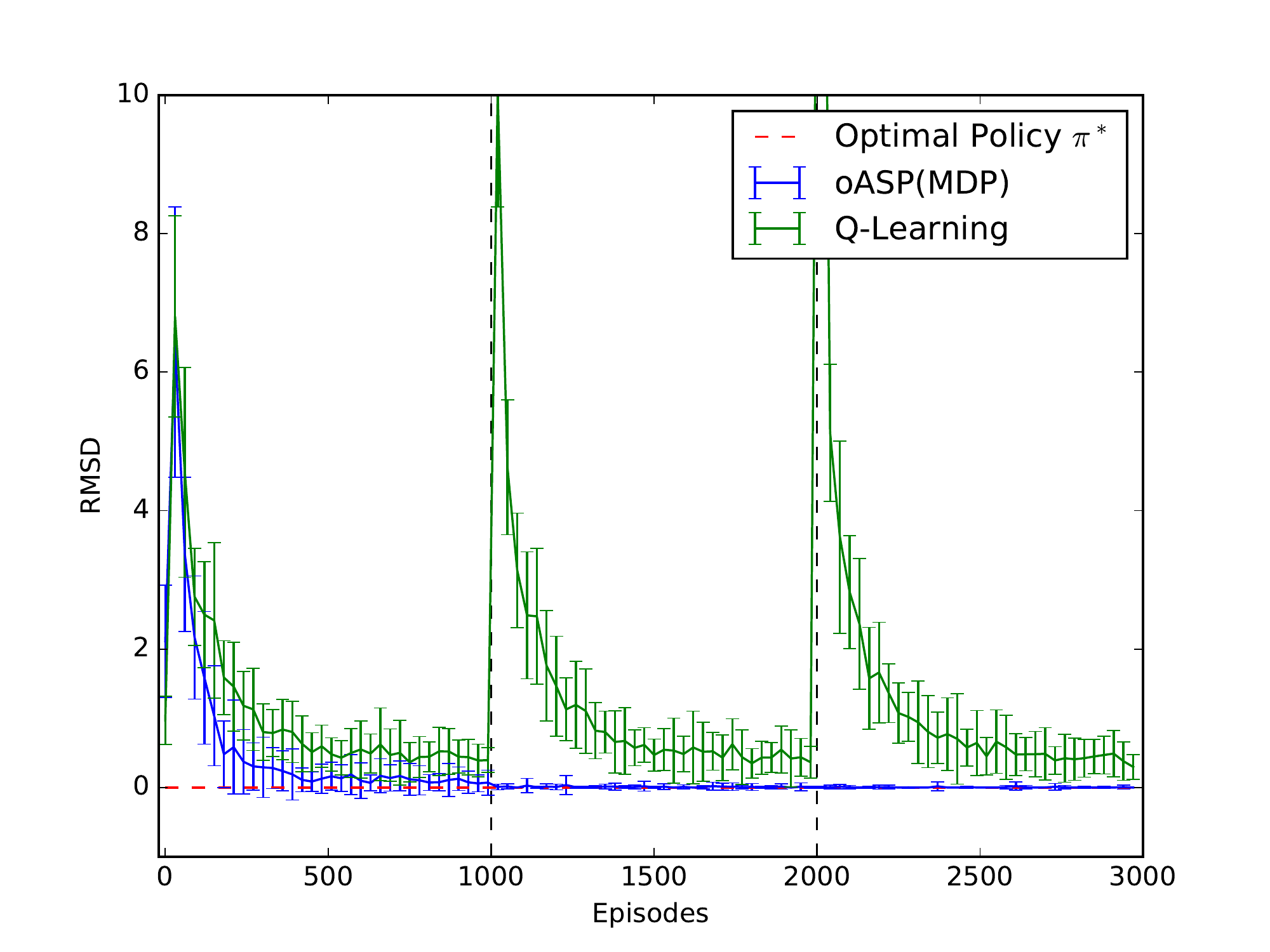}
    \caption{RMSD.}
	\label{fig:rmsd1z}
\end{subfigure}
\begin{subfigure}[t]{0.85\columnwidth}
	\centering
	\includegraphics[width=\columnwidth]{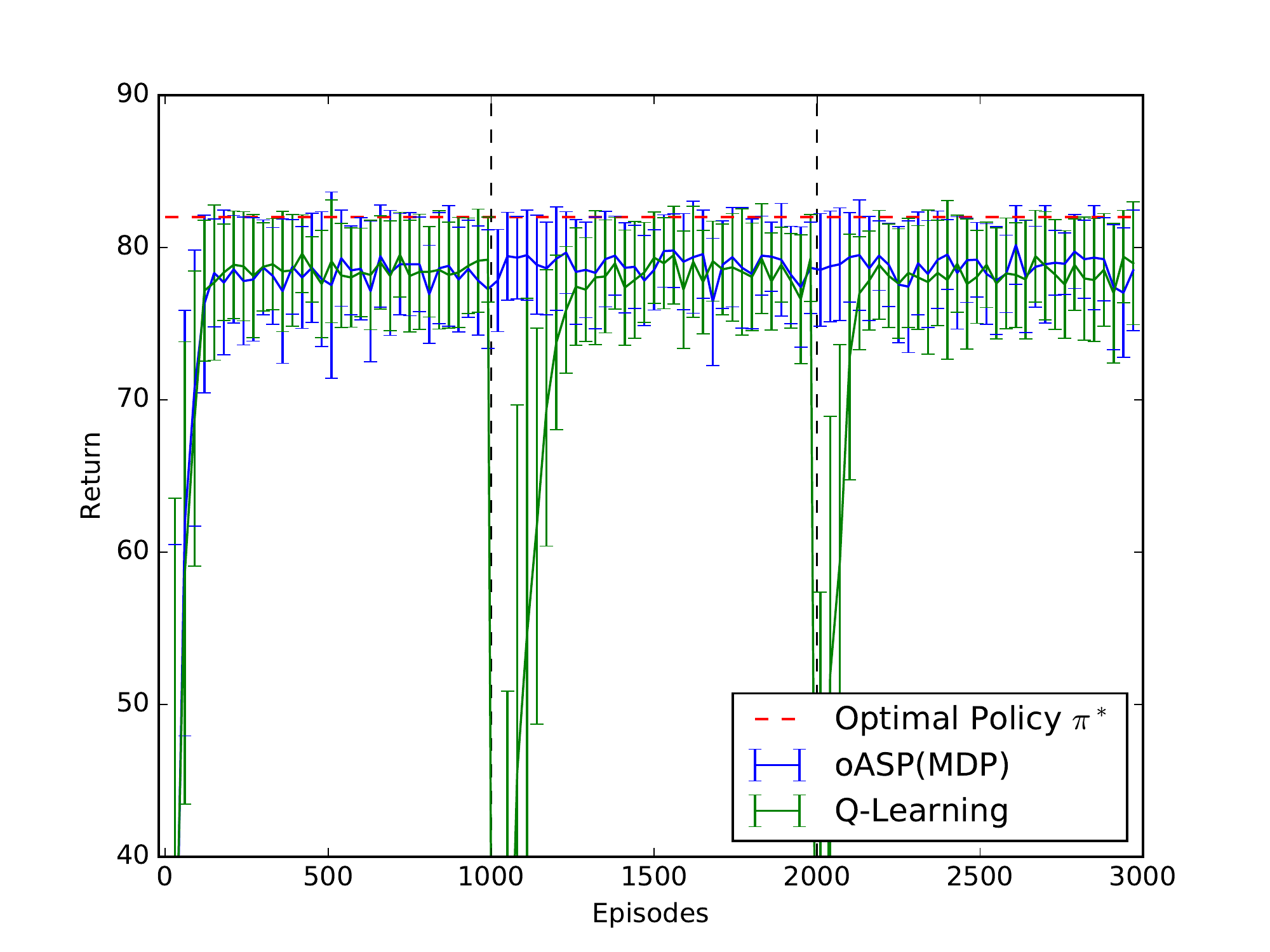}
    \caption{Return.}
	\label{fig:ret1z}
\end{subfigure}
\\
\begin{subfigure}[t]{0.85\columnwidth}
	\centering
	\includegraphics[width=\columnwidth]{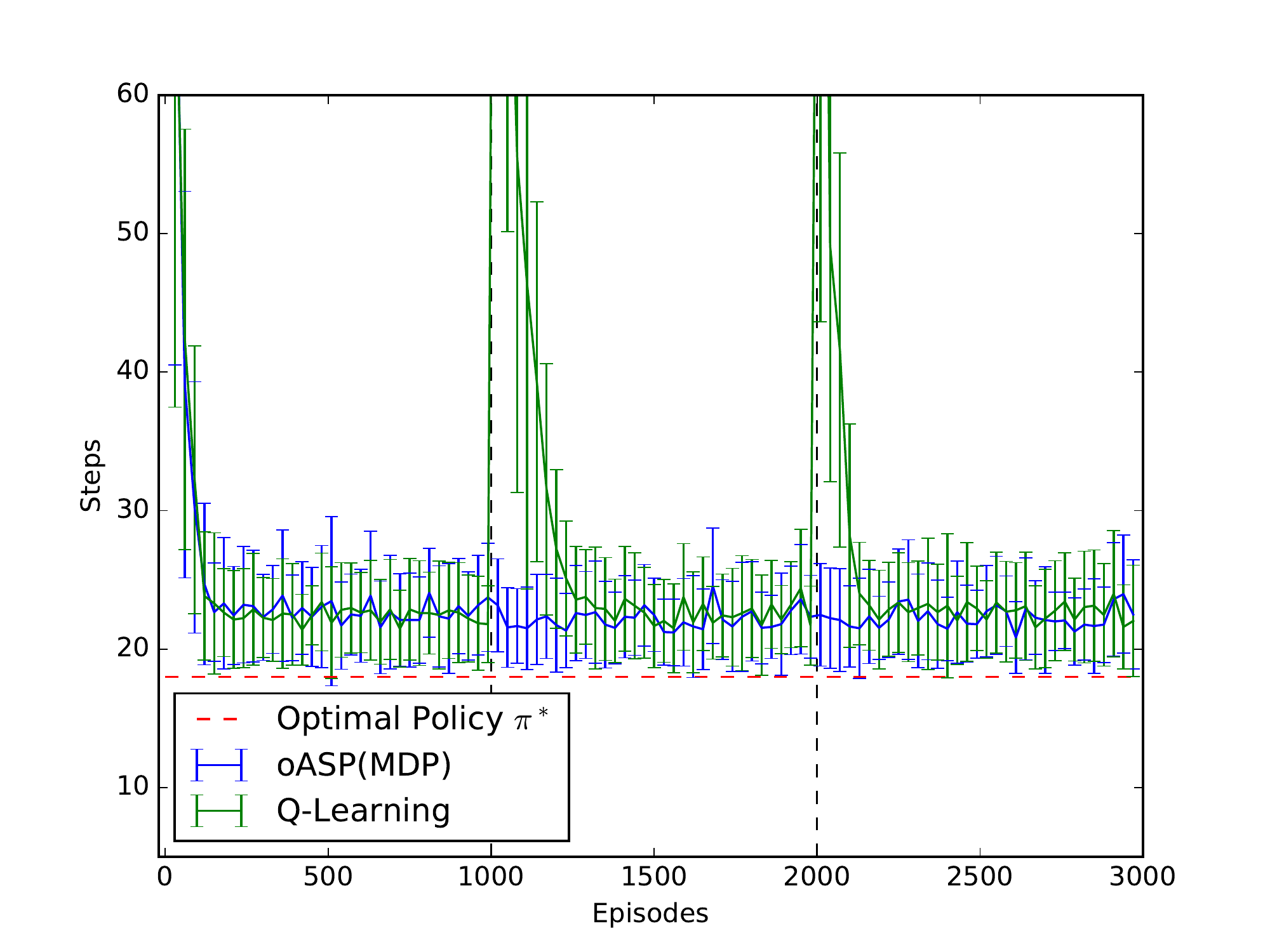}
    \caption{Number of steps.}
	\label{fig:steps1z}
\end{subfigure}
\begin{subfigure}[t]{0.75\columnwidth}
	\centering
	\includegraphics[width=\columnwidth]{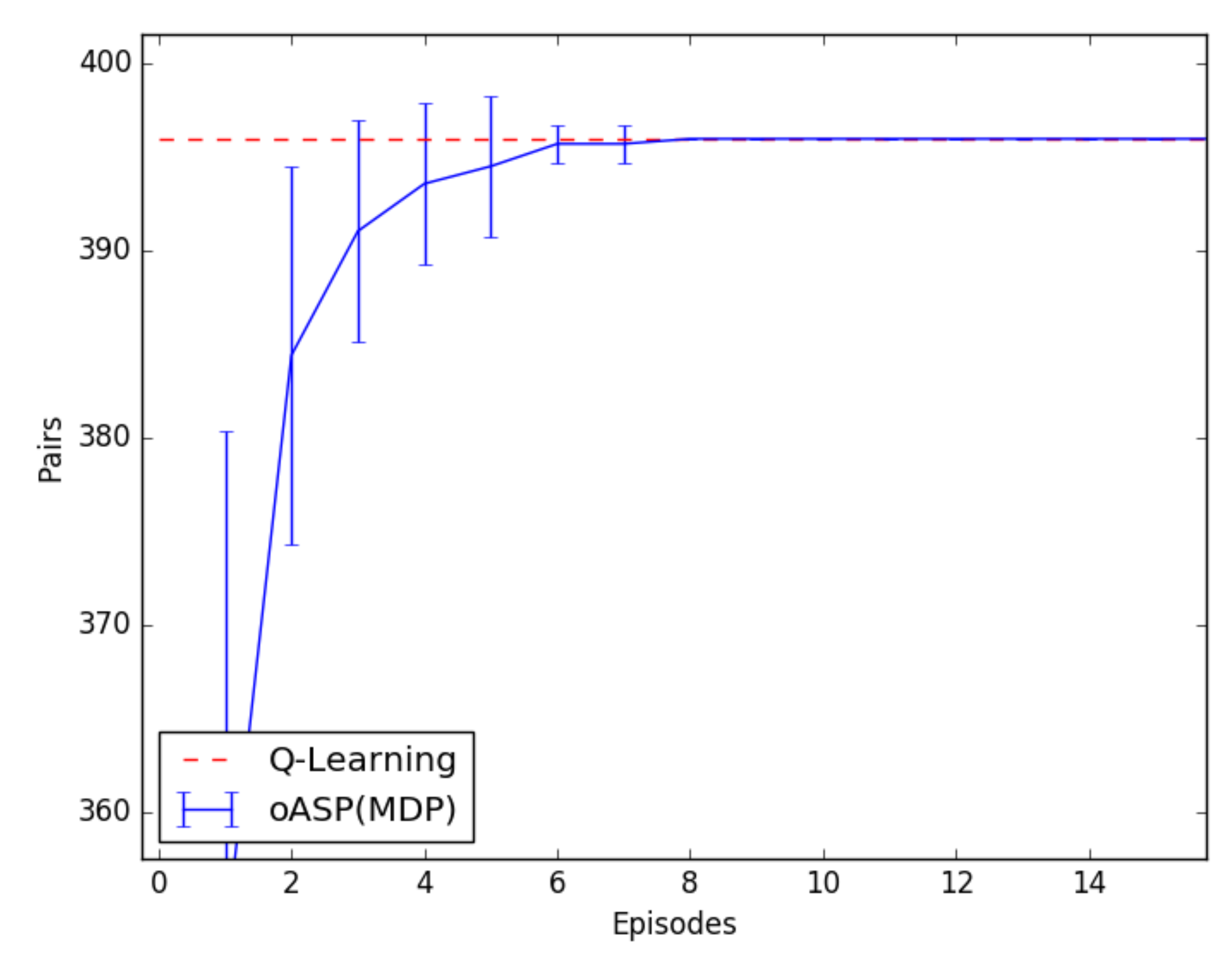}
	\caption{Numbers of state-action pairs.}
	\label{fig:pairs1c}
\end{subfigure}
\caption{Results for the first test.}
\label{fig:rexp1}
\end{figure}

The results obtained in the first test are represented in Figure~\ref{fig:rexp1}. Figure~\ref{fig:rmsd1z} shows that the RMSD values of oASP(MDP) decrease faster than those of Q-Learning, thus converging to the optimal policy ahead of Q-Learning. It is worth observing that when a change occurs in the environment (at episodes 1000 and at 2000) there is no increase in oASP(MDP) RMSD values, contrasting with the significant increase in Q-Learning's values. A similar behaviour is shown in Figure~\ref{fig:steps1z}, where there is no change in the number of steps of oASP(MDP) after a  change occurs, at the same time that Q-learning number of steps increase considerably at that point. 

The return values obtained in this test are shown in Figure~\ref{fig:ret1z}, where it can be observed that both oASP(MDP) and Q-learning reach the maximum value together during the initial episodes,  but there is no reduction in the return values of oASP(MDP) when the environment changes, whereas Q-learning returns drop to the initial figures. 

Figure~\ref{fig:pairs1c} shows the number of state-action pairs that oASP(MDP) has found for the grid world. Values obtained after the 15$^{th}$ episode were omitted since they presented no variation. This figure shows that oASP(MDP) has explored every state of the grid world and performed every action allowed in each state, resulting in a complete description of the environment. Since oASP(MDP) has provided the complete description of the environment, the agent that uses oASP(MDP) optimizes the same action-value function as the agent that uses Q-Learning, thus the optimal policy found by both agents is the same. Due to the exploration of the environment performed in the beginning of the interaction, before the 10\(^{th}\) episode the agent has executed every action in every possible state at least once and, as can be seen in line~\ref{alg:ifline} of Algorithm~\ref{alg:oaspmdp}, the agent then uses the underlying RL procedure to find the action-value function.


\subsection{Second test: changes in the transition probabilities}\label{sec:exp3}
In this test, the grid was fixed at a 10$\times$10 size, with wall--free-space ratio fixed at 25\%. Changes in the environment occurred with respect to the transition probabilities. Initially, the agent's actions had 50\% of probability for moving the agent in the desired direction and 25\% for moving it in each of the two orthogonal directions. The first change set the probabilities at 75\% (assigned to the desired action effect) and 12.5\% (for the directions orthogonal to the desired). The final change assigned 90\% for moving in the desired direction and 5\% for moving in each of the orthogonal directions.

\begin{figure}[ht!]
\centering
\begin{subfigure}[t]{0.85\columnwidth}
	\centering
	\includegraphics[width=\columnwidth]{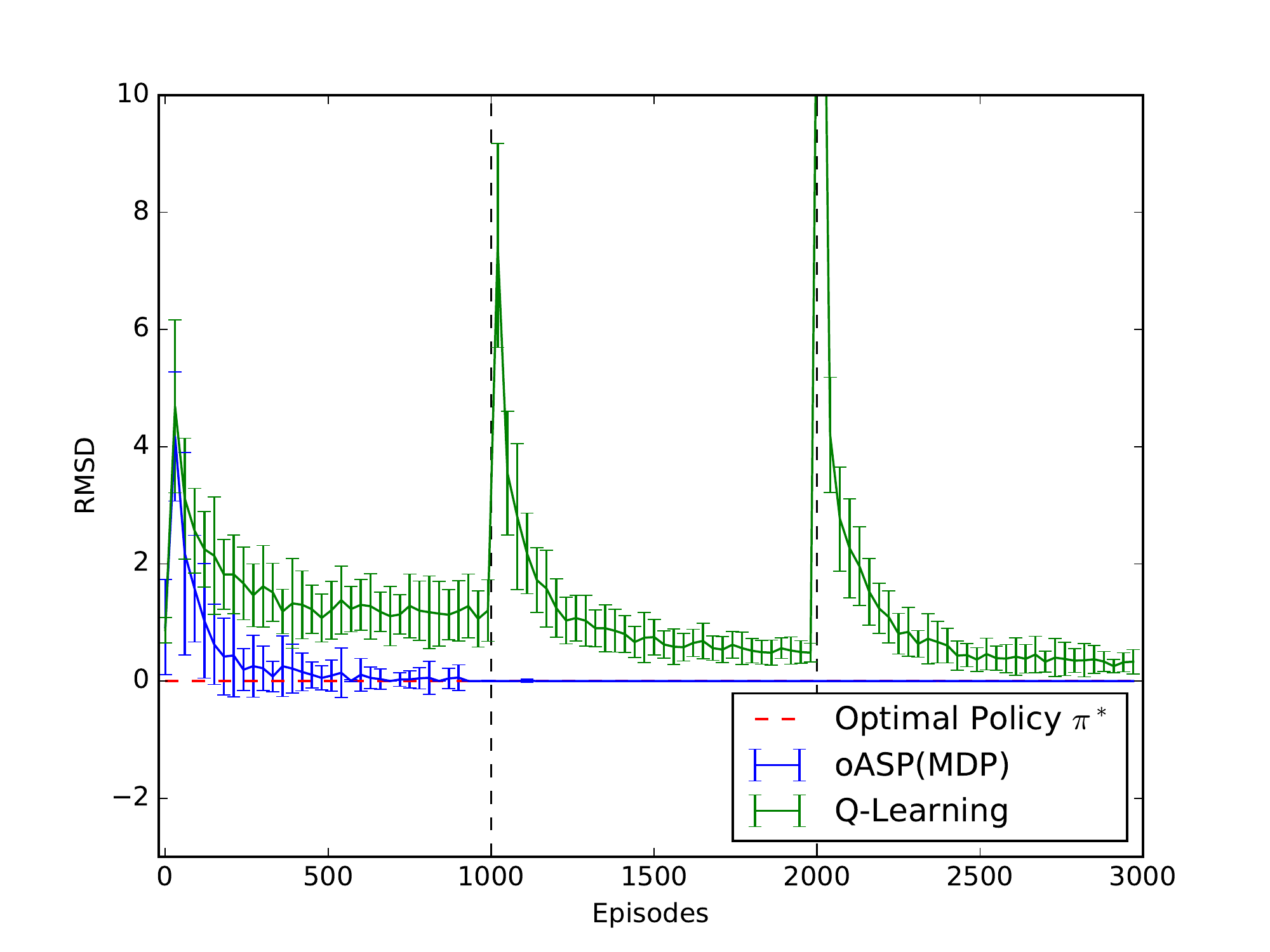}
    \caption{RMSD.}
	\label{fig:rmsd3z}
\end{subfigure}
\begin{subfigure}[t]{0.85\columnwidth}
	\centering
	\includegraphics[width=\columnwidth]{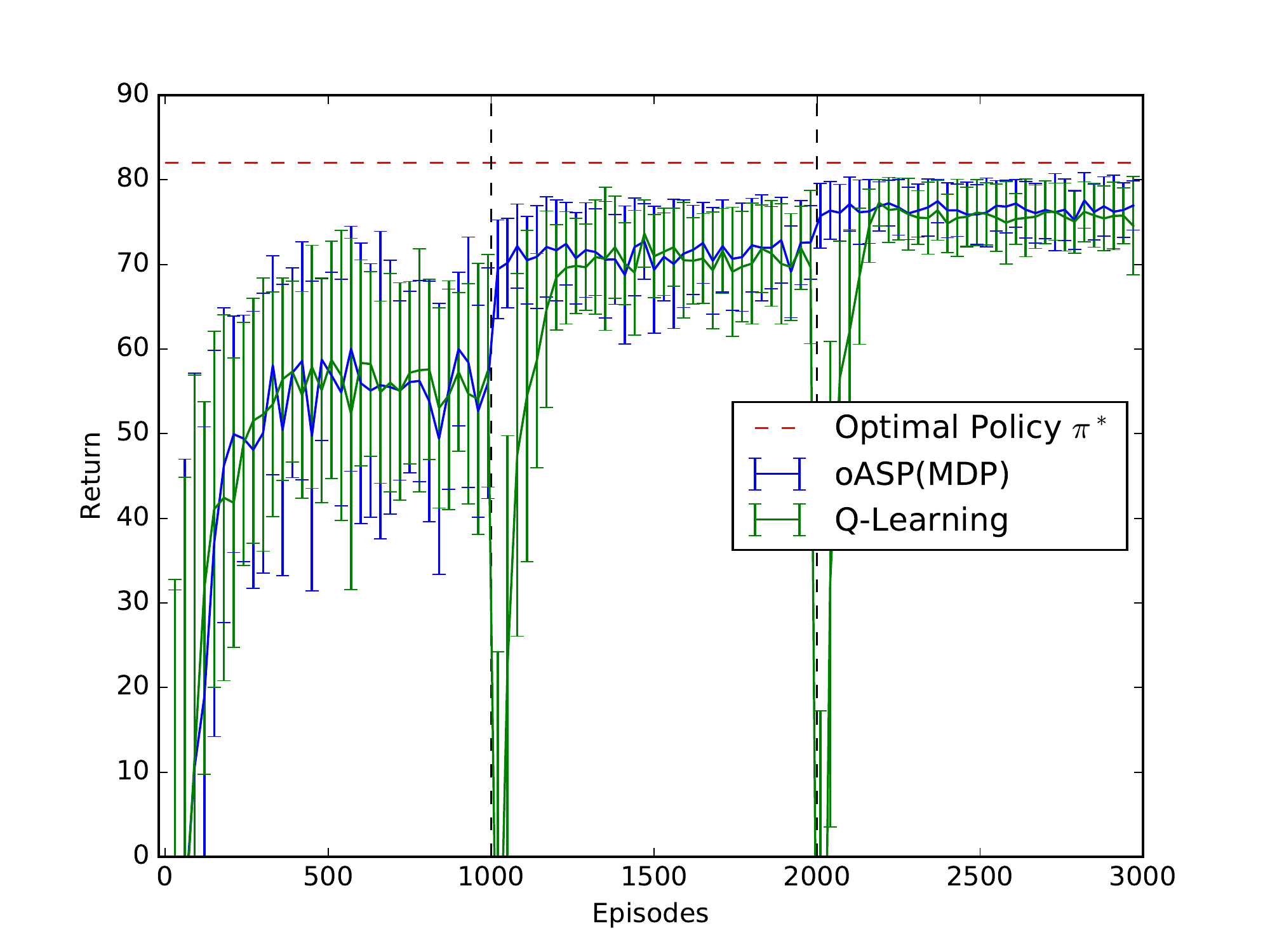}
    \caption{Return.}
	\label{fig:ret3z}
\end{subfigure}
\begin{subfigure}[t]{0.85\columnwidth}
	\centering
	\includegraphics[width=\columnwidth]{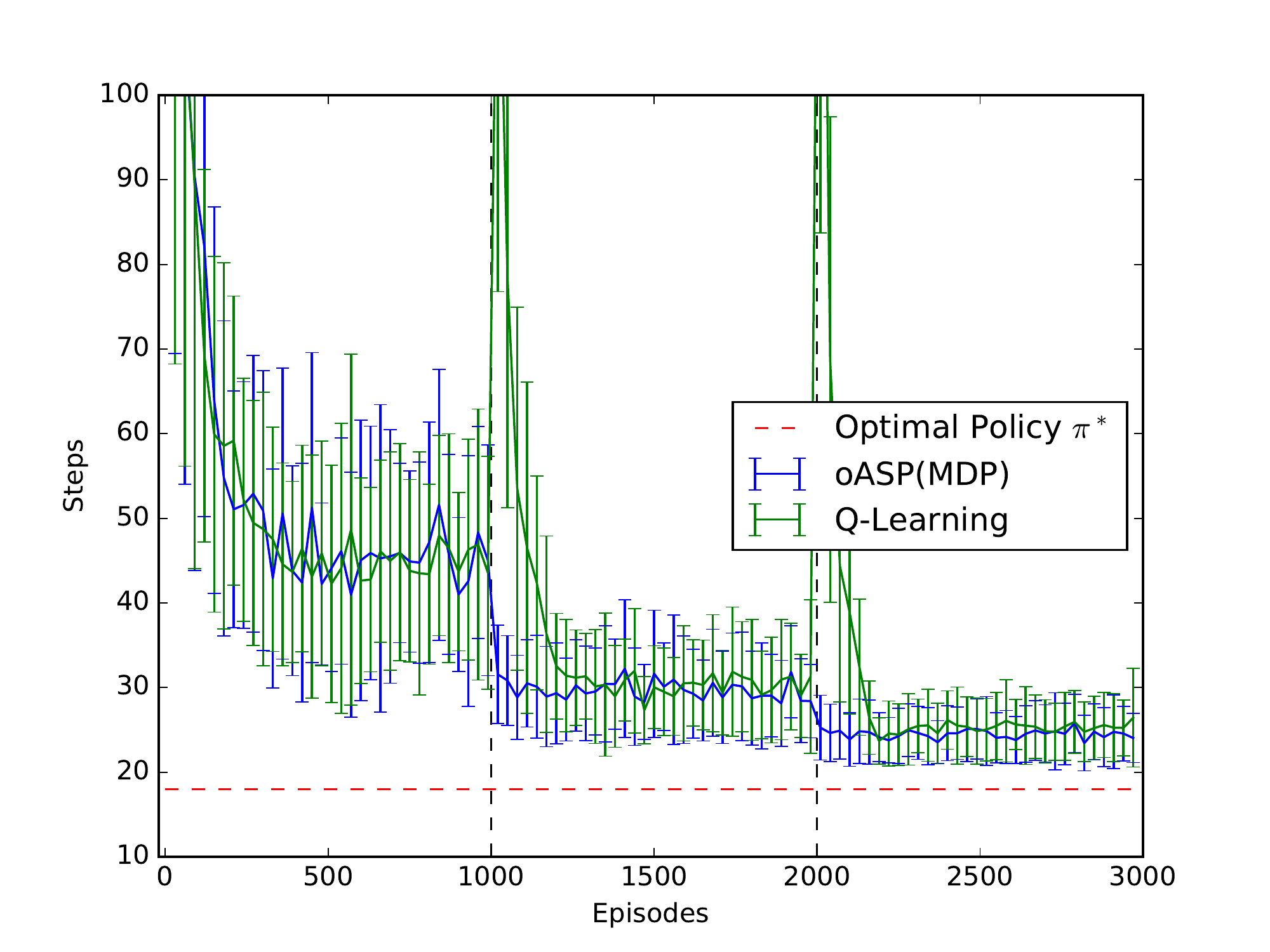}
    \caption{Number of steps.}
	\label{fig:steps3z}
\end{subfigure}
\begin{subfigure}[t]{0.75\columnwidth}
	\centering
	\includegraphics[width=\columnwidth]{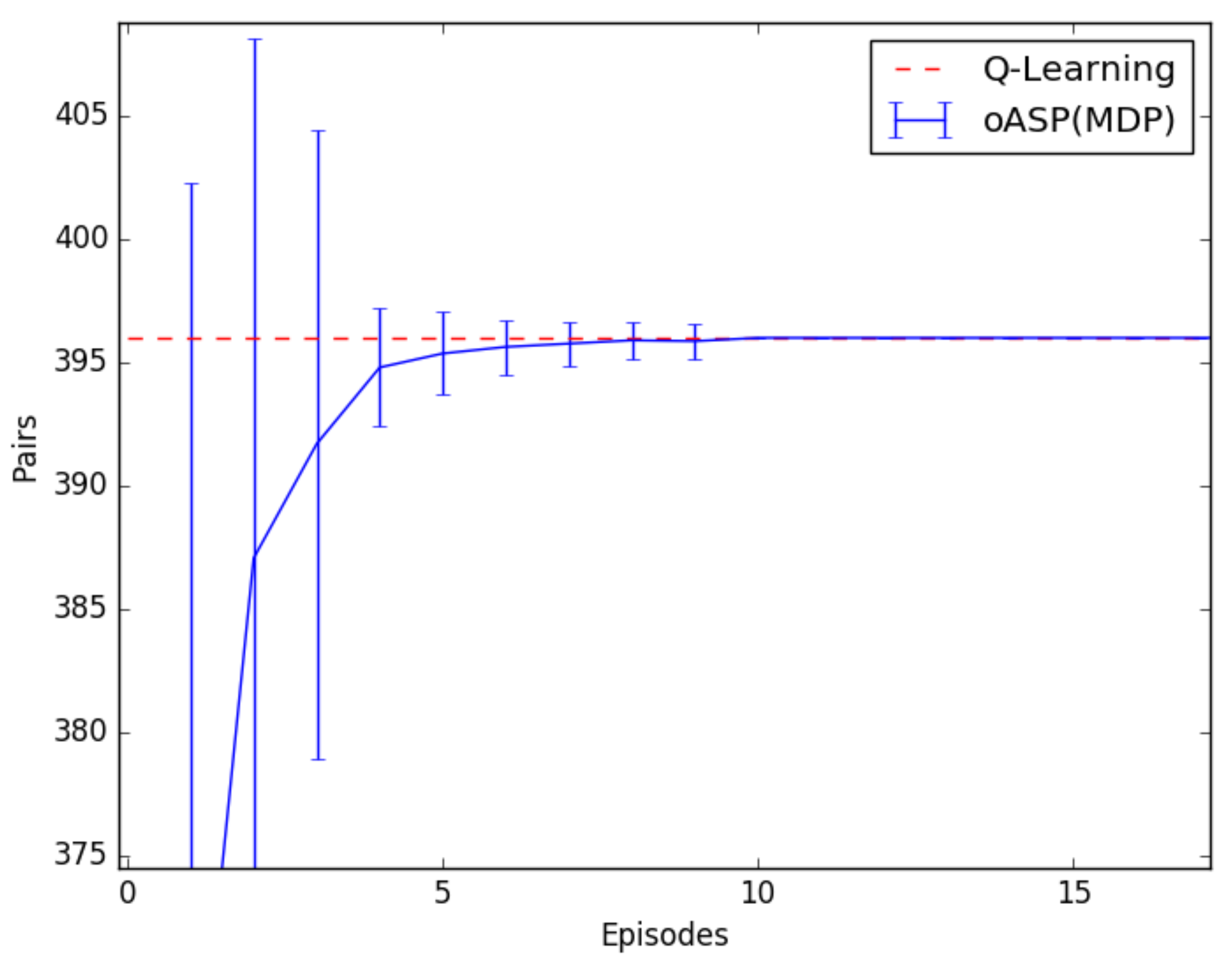}
	\caption{Number of state-action pairs.}
	\label{fig:pairs3c}
\end{subfigure}
\caption{Results for the second test.}
\label{fig:rexp3}
\end{figure}

The RMSD values for oASP(MDP), in this case, decreased faster than those of Q-Learning, reaching  zero before the first change occurred, while Q-Learning at that point had not yet converged, as shown in Figure \ref{fig:rmsd3z}. Analogously to the first test, there is no change in RMSD values of oASP(MDP) when the environment changes, whereas Q-learning presents re-initializations. In the results on return and the number of steps, shown in Figures~\ref{fig:ret3z} and~\ref{fig:steps3z} respectively, the performance of oASP(MDP) improves faster than the Q-Learning performance when there is a change in the environment. This is explained by the fact that, after oASP(MDP) approximates the action-value function (in the periods between the changes), when a change occurs, the information about it, acquired by the agent, is used to find solutions in the new world situation. In this case, the current action-value function is simply updated. Q-Learning, on the other hand, is restarted at each time a change occurs, resulting in the application of an inefficient policy in the new environment.

The number of state-action pairs that oASP(MDP) was able to describe is shown in Figure~\ref{fig:pairs3c}. Once more, values obtained after the 15\(^{th}\) episode were omitted, as they present no variation after this point. Analogous to the results obtained in the first experiment, oASP(MDP) was capable of executing at least once every allowed action in every state possible to be visited. As before, by exploring the environment oASP(MDP) could efficiently find the set of allowed states, defining the complete \(Q(s,a)\). 


In summary, the tests performed in the domains considered show that the information previously obtained is beneficial to an agent that learns by interacting with a changing environment. The action-value function obtained by oASP(MDP) before a change occurs accelerates the approximation of this function in a new version of the environment, avoiding the various re-initializations observed in Q-learning alone  (as shown in Figures \ref{fig:rexp1} and \ref{fig:rexp3}). However, as the action-value function approximation method used in oASP(MDP) (in this work) was Q-Learning, the policies learnt by oASP(MDP) and Q-Learning alone were analogous. This can be observed when comparing the curves for oASP(MDP) and Q-Learning in Figures \ref{fig:rexp1} and \ref{fig:rexp3} after convergence, noticing also that they keep the same distance with respect to the best performance possible (red-dashed lines in the graphs).



Tests were performed in virtual machines in AWS EC2 with t2.micro configuration, which provides one virtual core of an Intel Xeon at 2.4GHz, 1GB of RAM and 8GB of SSD with standard Debian 8 (Jessie). oASP(MDP) was implemented in Python 3.4 using ZeroMQ for providing messages exchanges between agent and environment and Clingo \cite{ASP2012} was used as the ASP Engine. The source code for the tests can be found in the following (anonymous) URL: \url{http://bit.ly/2k03lkl}.
\section{Related Work}
Previous attempts at combining RL with ASP include \cite{Mohan2015IEEE}, which proposes the use of ASP to find a pre-defined plan for a RL agent. This plan is described as a hierarchical MDP and RL is used to find the optimal policy for this MDP. However, changes in the environment, as used in the present work, were not considered in \cite{Mohan2015IEEE}.

Analogous methods were proposed by \cite{khandelwal2014,yang2014}, in which an agent interacts with an environment and updates an action's cost function. While \cite{khandelwal2014} uses the action language \(BC\), \cite{yang2014} uses ASP to find a description of the environment. Although both methods consider action costs, none of them uses Reinforcement Learning and they do not deal with changes in the action-value function description during the agent's interaction with the environment.

An approach to non-deterministic answer set programs is P-Log \cite{Baral-PLog,p-log}. While P-Log is capable of calculating transition probabilities from sampling, it is not capable of using this information to generate policies. Also P-Log  does not consider action costs. Thus, although P-Log  can be used to find the transition function, it cannot find the optimal solution, as proposed here.

Works related to non-stationary MDPs such as \cite{oMDP,ArbRew}, which deal only with changes in reward function, are more associated with RL alone than with a hybrid method such as oASP(MDP), since RL methods are already capable of handling changes in the reward and transition functions. The advantage of ASP is to find the set of states so that it is possible to search for an optimal solution regardless of the agent's transition and reward functions.

A proposal that closely resembles oASP(MDP) is \cite{Shanahan2016}. This method proposes the combination of deep learning to find a description to a set of states, which are then described as rules to a probabilistic logic program and, finally, a RL agent interacts with the environment using the results and learns the optimal policy. 
\section{Conclusion}\label{sec:conc}
This paper presented the method oASP(MDP) for approximating action-value functions of Markov Decision Processes, in non-stationary domains, with unknown set of states and unknown transition and reward functions. This method is defined on a combination of Reinforcement Learning (RL) and Answer Set Programming (ASP). The main advantage of RL is that it does not need \emph{a priori} knowledge of transition and reward functions, but it relies on having a complete knowledge to the set of domain states. In oASP(MDP), ASP is used to construct the set of states of an MDP to be used by a RL algorithm. ASP programs representing domain states and transitions are obtained as the agent interacts with the environment. This provides an efficient solution to finding optimal policies in changing environments.

Tests were performed in two non-stationary non-deterministic grid-world domains, where each domain had one property of the grid world changed over time. In the first domain, the ratio of obstacles and free space in the grid was changed, whereas in the second domain changes occurred in the transition probabilities. The changes happened in intervals of 1000 episodes in both domains. Results show that, when a change occurs, oASP(MDP) (with Q-learning as the action-value function) is capable of approximating the \(Q(s,a)\) function faster than Q-learning alone. Therefore, the combination of ASP with RL was effective in the definition of a method that provides more general (or more elaboration tolerant) solutions to changing domains than RL methods alone. 


Future work will be directed toward the development of an interface to facilitate the use of oASP(MDP) with distinct domains, such as those provided by the DeepMind Lab \cite{DeepMindLab}. Also, a comparison of oASP(MDP) with the framework proposed in \cite{Shanahan2016} is an interesting subject for future research.



\clearpage
\bibliographystyle{named}
\bibliography{references}

\end{document}